\documentclass[runningheads]{llncs}

\usepackage{XLoKR-2022-clustering}

\begin{document}

\title{
	Clustering-Based Approaches for Symbolic Knowledge Extraction
}

\author{
	Federico Sabbatini\thanks{Corresponding author}\inst{1}\orcidID{0000-0002-0532-6777}
	\and \\
	Roberta Calegari\inst{2}\orcidID{0000-0003-3794-2942}
}

\authorrunning{F.\ Sabbatini et al.}

\institute{
	Dipartimento di Scienze Pure e Applicate (DiSPeA), \\Universit\`a degli Studi di Urbino ``Carlo Bo'', Italy\\ \email{f.sabbatini1@campus.uniurb.it}
	\and
	Alma AI---Alma Mater Research Institute for Human-Centered Artificial Intelligence, \\\textsc{Alma Mater Studiorum}--Universit\`a di Bologna, Italy\\ \email{roberta.calegari@unibo.it}
}

\maketitle

\begin{abstract}

Opaque models belonging to the machine learning world are ever more exploited in the most different application areas.
These models, acting as black boxes (BB) from the human perspective, cannot be entirely trusted if the application is critical unless there exists a method to extract symbolic and human-readable knowledge out of them.

In this paper we analyse a recurrent design adopted by symbolic knowledge extractors for BB regressors---that is, the creation of rules associated with hypercubic input space regions.
We argue that this kind of partitioning may lead to suboptimal solutions when the data set at hand is high-dimensional or does not satisfy symmetric constraints.
We then propose a (deep) clustering-based approach to be performed before symbolic knowledge extraction to achieve better performance with data sets of any kind.


\end{abstract}

\section{Introduction}

Machine learning (ML) models in general -- and (deep) artificial neural networks in particular -- are nowadays exploited to draw predictions in almost every application area~\cite{rocha2012far}.
However, when facing \emph{critical} domains -- e.g., involving human health or wealth -- ML models behaving as opaque predictors are not an acceptable choice.
The \emph{opaque} nature of these models makes them unintelligible for humans and this is the reason why they are called \emph{black boxes} (BBs).
%
%
Nonetheless, explainability can be obtained from BBs via several strategies~\cite{guidotti2018survey}.
For instance, one can rely uniquely on \emph{interpretable} models~\cite{Rudin2019}, or build explanations by applying reverse engineering to the BB behaviour~\cite{KENNY2021103459}.
This latter approach allows users to combine the impressive predictive capabilities of opaque models with the human readability proper of symbolic models.

The literature offers a wide variety of procedures explicitly designed to extract symbolic knowledge from opaque ML models, especially for classifiers -- e.g., Rule-extraction-as-learning~\cite{craven1994using} and \trepan{}~\cite{craven1996extracting} --, but also for BB regressors---e.g., \iter{}~\cite{huysmans2006iter}, \gridex{}~\cite{gridex-extraamas2021}, \gridrex{}~\cite{gridrex-kr2022} and \textsc{RefAnn}~\cite{setiono2002extraction}.

Unfortunately, any method is subject to drawbacks and limitations.
In the following we focus on the issues deriving from the extraction of rules from BB regressors.
In particular, we observe that a typical design choice adopted for extractors applicable to opaque regressors is to find hypercubic regions in the input feature space having similar instances and then to associate symbolic knowledge to each region, for instance in the form of \emph{first-order logic} rules~\cite{huysmans2006iter,gridrex-kr2022,gridex-extraamas2021}.
We agree that rules associated with hypercubic regions are the best choice from the human-readability perspective, since they enable the description of an input space region in terms of constraints on single dimensions (e.g., $0.3 < X < 0.6, 0.5 < Y < 0.75$ for a hypercube in a 2-dimensional space having features $X$ and $Y$).
However this solution may lead to the creation of suboptimal clusters, especially if the partitioning into hypercubes is performed following some sort of symmetric procedure (on asymmetric data sets).
In this paper we suggest to exploit clustering techniques on the data set used to train the BB before extracting knowledge from it, in order to preemptively find and distinguish relevant input regions with the corresponding boundaries.
This theoretically allows extractors to:
\begin{inlinelist}
	\item automatically tune the number of output rules w.r.t.\ the number of relevant regions found;
	\item give priority to more relevant regions, for instance those containing more input instances or having the largest volume;
	\item avoid unsupervised partitioning of the input feature space, otherwise leading to suboptimal solutions in terms of readability and/or fidelity.
\end{inlinelist}

\section{Related Works}\label{sec:state}

A predictive model can be defined as \emph{interpretable} if human users are able to easily understand its behaviour and outputs~\cite{agentbasedxai-aamas2020}.
Since the majority of modern ML predictors store the knowledge acquired during their training phase in a \emph{sub-symbolic} way, they behave and appear to the human perspective as unintelligible black boxes.
The XAI community has proposed a variety of methods to enrich BB predictions with corresponding interpretations/explanations without renouncing their superior predictive performance.
Usually, the proposed methods consist of creating an interpretable, mimicking model by inspecting the underlying BB in terms of internal behaviour and/or input/output relationships.
For instance, \textsc{RefAnn} analyses the architecture of neural network regressors with one hidden layer to obtain information about the internal parameters and thus build human-readable \emph{if-then} rules having a linear combination of the input features as postconditions.
This kind of technique is called \emph{decompositional}.
On the other hand, when the internal structure of the BB is not considered to build explanations, the algorithms are classified as \emph{pedagogical}.
In the following we focus on three pedagogical techniques for BB regressors---namely, \iter{}, \gridex{} and \gridrex{}.

\paragraph{\iter{}}

The \iter{} algorithm~\cite{huysmans2006iter} is a pedagogical technique to extract symbolic knowledge from BB regressors.
It follows a bottom-up strategy, starting from the creation of infinitesimally small hypercubes in the input feature space and iteratively expanding them until the whole space is covered or it is not possible to add or expand further the existing cubes.
All the hypercubes are non-overlapping and they do not exceed the input feature space.
After the expansion step, \iter{} associates an \emph{if-then} rule to each cube, selecting as action the mean output value of all the instances contained in the cube.
The algorithm's main advantage is to be capable of constructing hypercubes having different dimensions.
However, especially when dealing with high-dimensional data sets, it may present several criticalities related to the hypercube expansion.
In particular, at each iteration it is necessary to build temporary cubes around the existing cubes, two per input dimension (as shown in the bidimensional example reported in \Cref{fig:exp}), but only one temporary cube of a single existing cube is chosen as actual expansion.
This may lead to an enormous waste of computational time and resources due to the repetition of (the same) useless calculations, other than the possibility to exceed the maximum allowed iterations without having convergence.
The absence of convergence results in a non-exhaustive partitioning of the input feature space, which in turn implies the inability to predict output values of data samples belonging to regions that are not covered by a hypercube (i.e., there are no human-interpretable predictive rules associated with uncovered regions).
Conversely, the coverage of the whole input feature space enables to draw predictions for any input instance.

\newcommand\demodim{0.23\linewidth}

\begin{figure*}[tb]\centering
	\subfloat{
		\includegraphics[width=\demodim]{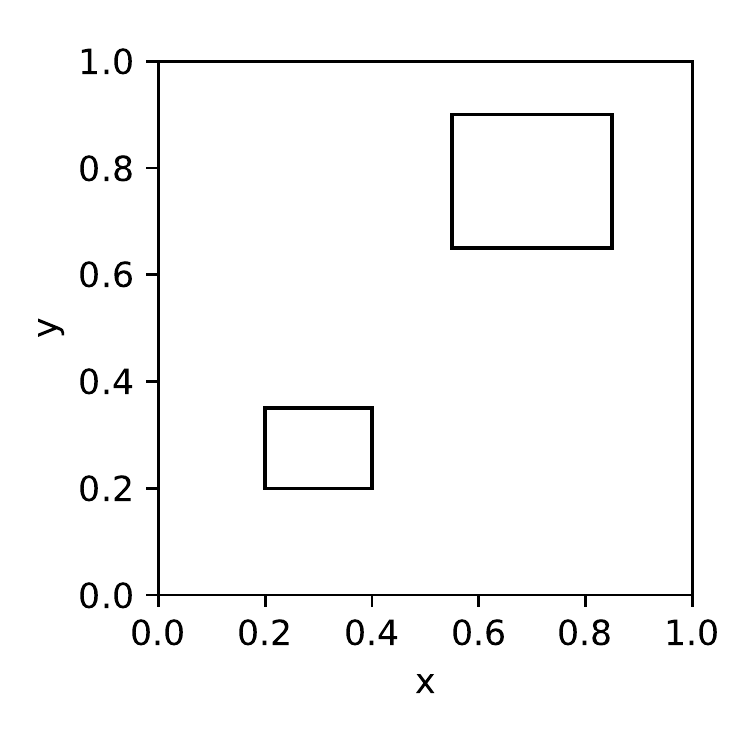}\label{fig:exp1}
	}
	\subfloat{
		\includegraphics[width=\demodim]{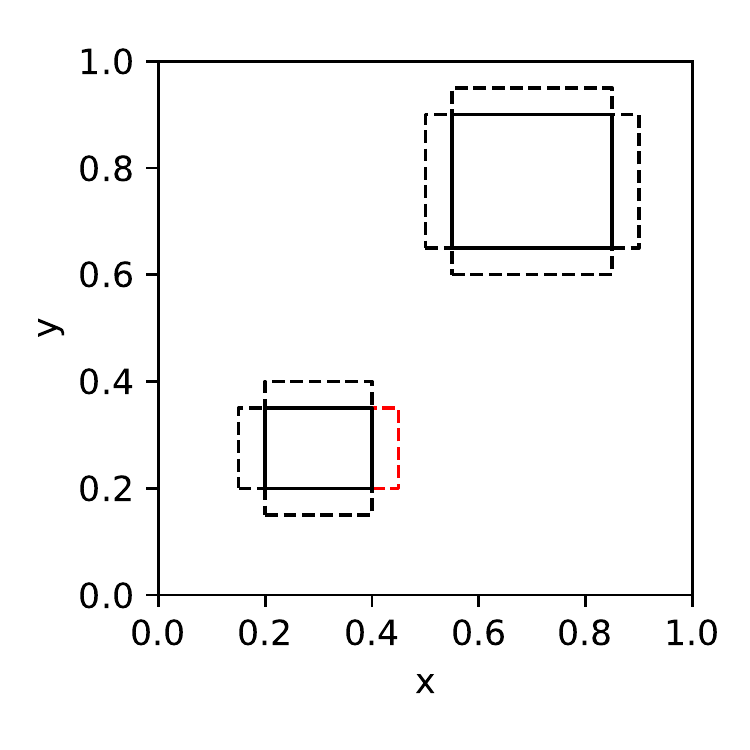}\label{fig:exp2}
	}
	\subfloat{
		\includegraphics[width=\demodim]{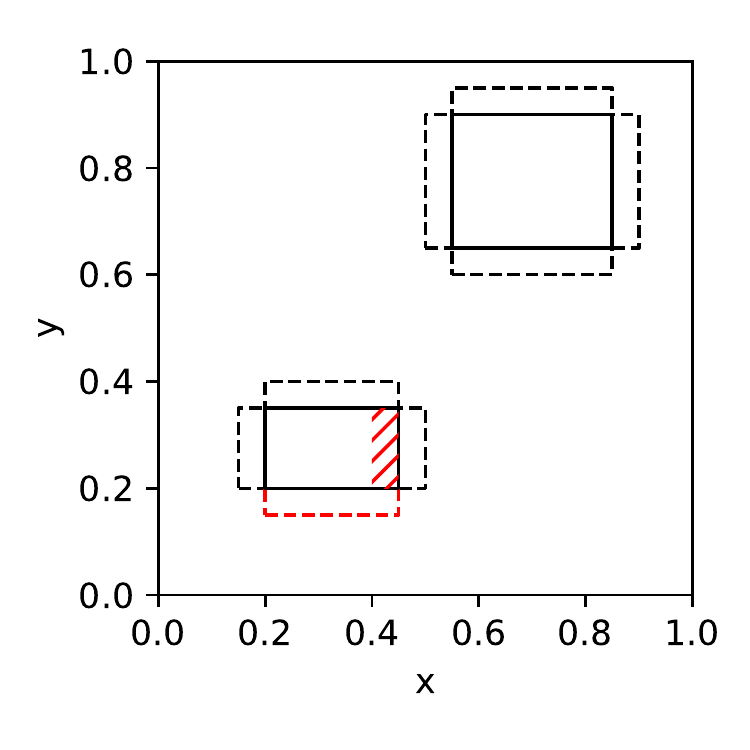}\label{fig:exp3}
	}
	\subfloat{
		\includegraphics[width=\demodim]{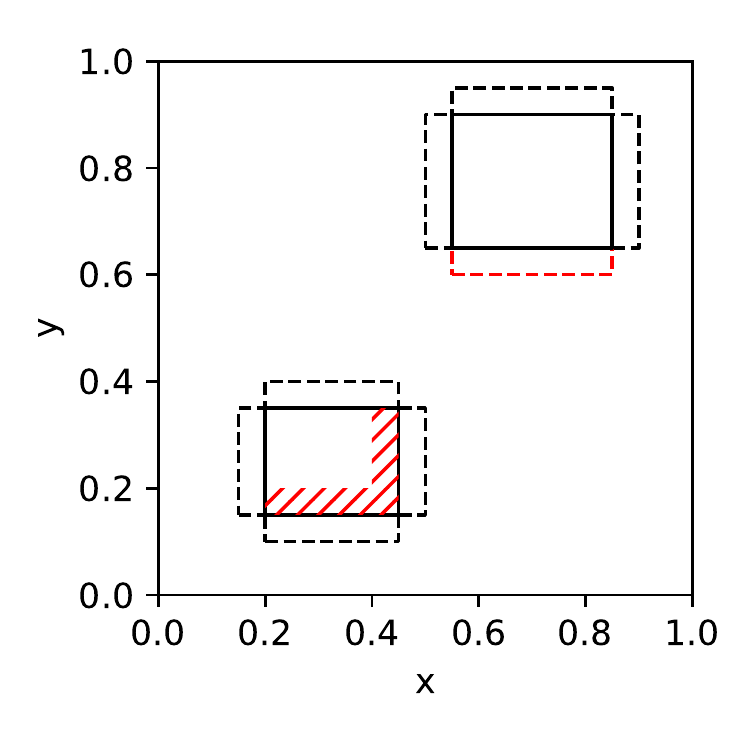}\label{fig:exp4}
	}
	\caption{Example of \iter{} hypercube expansion.}\label{fig:exp}
\end{figure*}

\paragraph{\gridex{} and \gridrex{}}

\begin{figure*}[tb]\centering
	\subfloat{
		\includegraphics[width=\demodim]{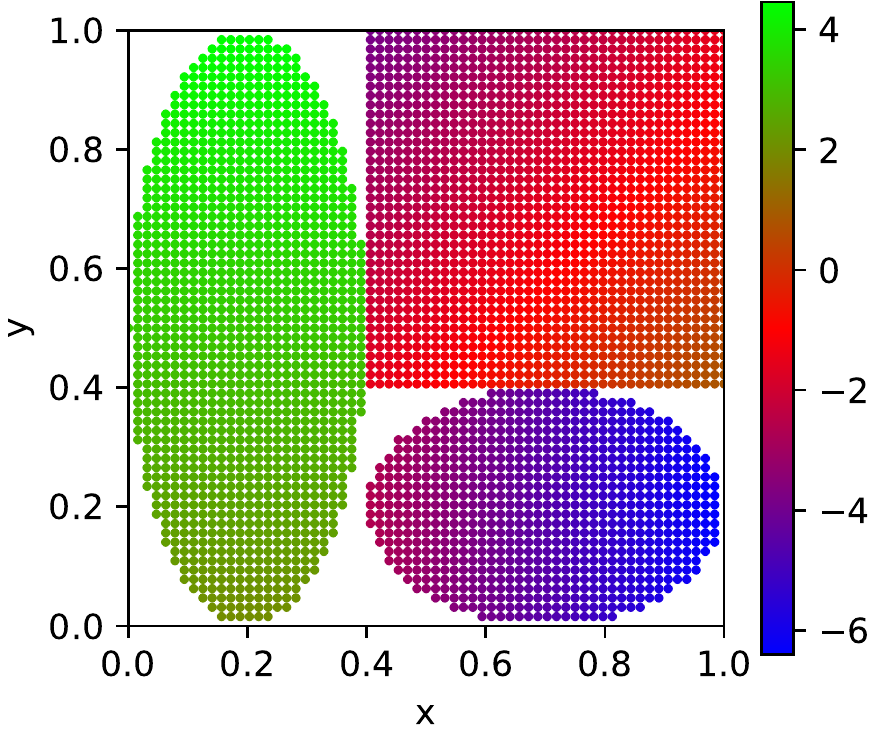}\label{fig:part1}
	}
	\subfloat{
		\includegraphics[width=\demodim]{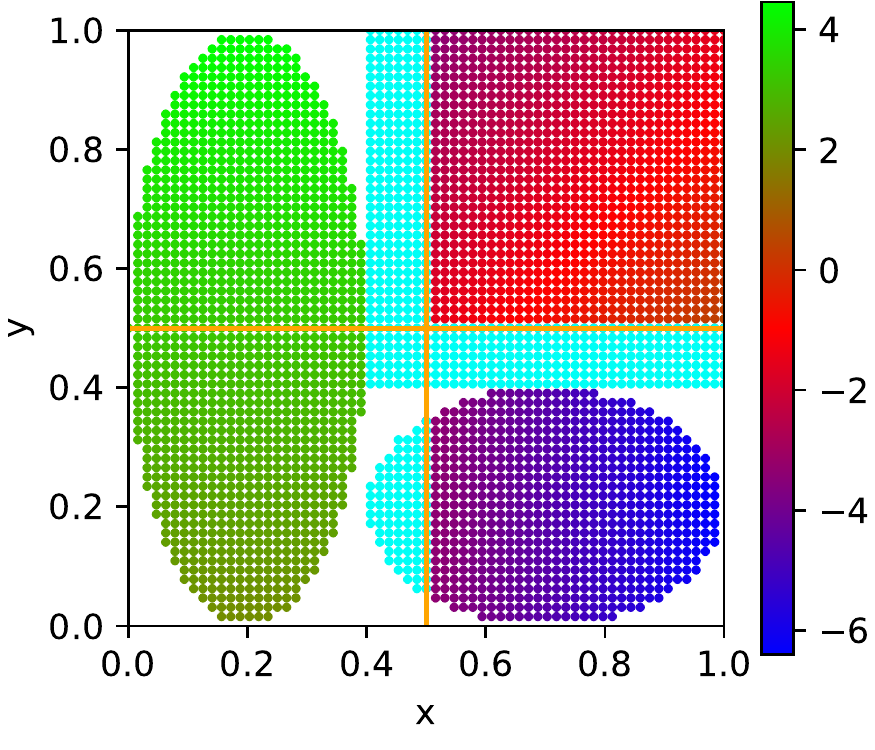}\label{fig:part2}
	}
	\subfloat{
		\includegraphics[width=\demodim]{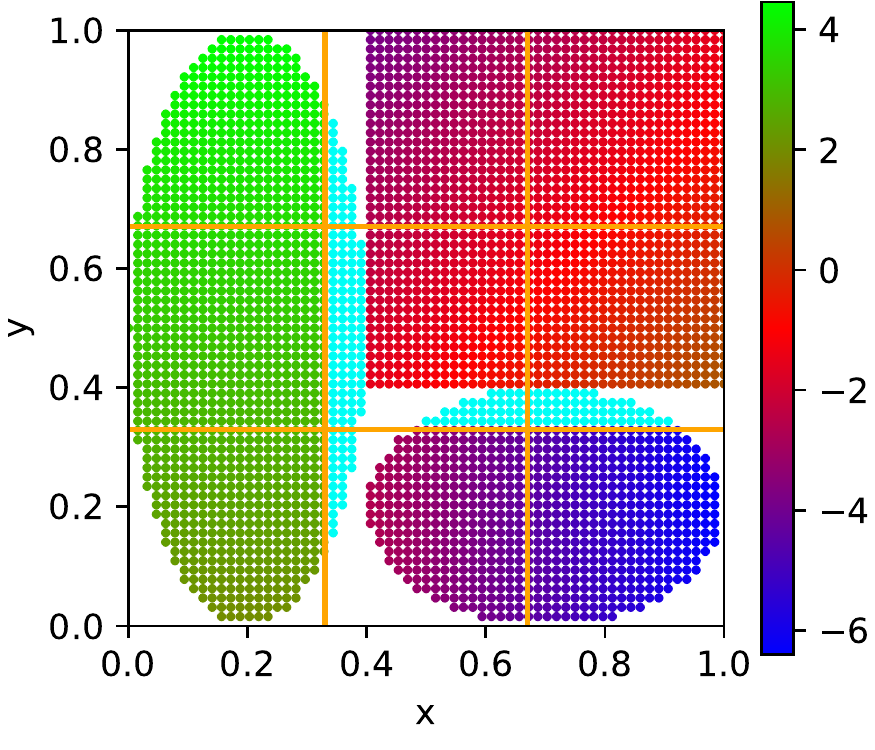}\label{fig:part3}
	}
	\subfloat{
		\includegraphics[width=\demodim]{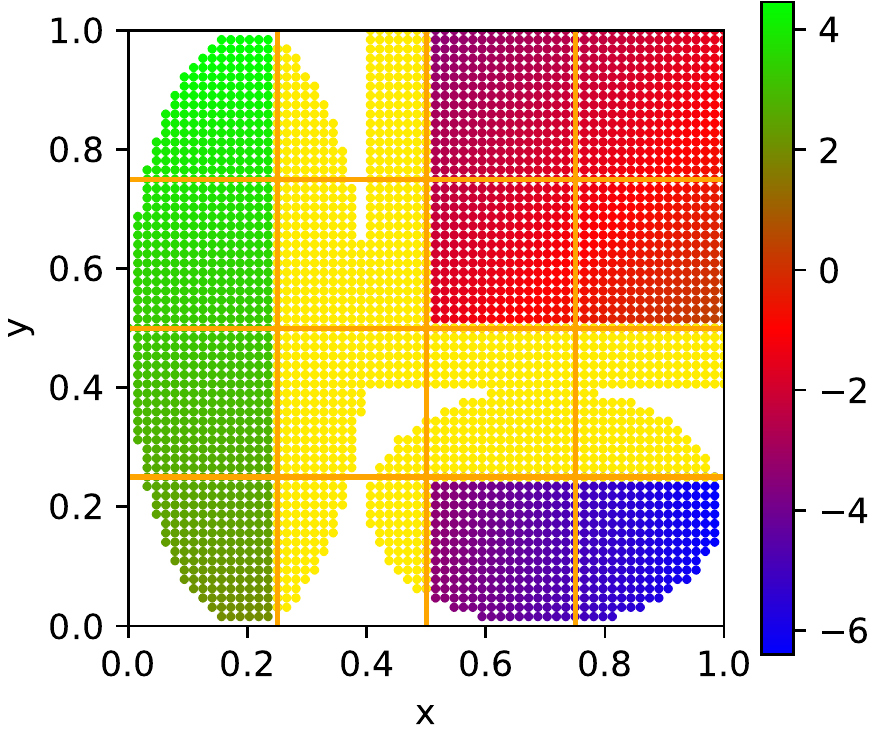}\label{fig:part4}
	}
	\caption{Example of \gridex{} partitioning (without merging step) when performing 2, 3 or 4 slices per input dimension. The first panel shows the input feature space. Cyan and yellow dots in the other panels represent small and large clustering errors, respectively.}\label{fig:part}
\end{figure*}

The \gridex{} algorithm~\cite{gridex-extraamas2021} is a different pedagogical technique to extract symbolic knowledge from BB regressors.
It is applicable under the same conditions of \iter{} and outputs the same kind of knowledge, but they differ in the strategy adopted during the input space partitioning.
\gridex{} is not a bottom-up algorithm; conversely, it follows a top-down strategy, starting from the whole input space and recursively partitioning it into smaller hypercubic regions, according to a user-defined threshold acting as a trade-off criterion between readability (in terms of the number of extracted rules) and fidelity of the output model (intended as its ability to mimic the underlying BB).
Amongst the advantages of \gridex{} there is its ability to automatically refine the found regions according to the provided threshold, as well as to perform a merging step after each split, when possible.
In particular, the merging step consists of the pairwise unification of adjacent hypercubes to reduce the number of output rules and it is based on the similarity between the samples included in each cube, to avoid a predictive performance worsening.
It is useful since a split may create an excessive amount of disjoint regions.
The main disadvantage of \gridex{} is that, even if it can perform an arbitrary amount of slices along each dimension at each iteration, all the partitions created in a dimension in a given iteration have the same size.
An example is reported in \Cref{fig:part}, showing that a cube may include portions of separated input regions, thus leading to a drop in the predictive performance.

The \gridrex{} algorithm~\cite{gridrex-kr2022} is an extension of \gridex{} achieving better predictive performance by providing as output linear combinations of the input variables instead of constant values.
However, it performs the same grid partitioning than \gridex{}.
As a consequence, it is susceptible to the same partitioning-related drawbacks.

\section{A Clustering-Based Approach}\label{sec:contribute}

The example in \Cref{fig:part} shows a data set described by 2 input features (x and y) and an output continuous feature (represented by the colour).
The data set can be divided into three clusters of instances.
Inside each cluster, the output feature is represented by a different linear combination of the input features.
An optimal extraction technique should be able to identify these clusters in a \emph{fast} and \emph{flawless} way, since in the example they are linearly separable and, furthermore, three hypercubes can enclose them without overlaps.
A bottom-up strategy like the one adopted by \iter{} may be suitable for the task, but in the general case -- in the n-dimensional domain -- convergence may result too slow.
It is possible to fasten the \iter{} convergence by acting on the algorithm parameters, but at the expense of a coarser partitioning.
This latter inconvenience is the same encountered by using \gridex{}, which induces an equally spaced grid.
If a grid cell contains no instances belonging to different clusters, the predictive error will be small.
Otherwise, it will be less or more large depending on the amount of contamination of each cell (reported in the Figure as cyan and yellow dots if it is small or large, respectively).

An optimal and fast partitioning can be obtained by:
\begin{inlinelist}
	\item applying a clustering technique to the data set to identify the different relevant regions;
	\item constructing minimal hypercubes to include the found regions, since they will be associated with human-readable rules;
	\item exploiting an extraction technique able to describe each hypercube in terms of input features.
\end{inlinelist}
Extraction techniques may surely benefit from cluster-aware partitioning methods.
Accuracy in the selection of different clusters and in the construction of enclosing hypercubes may enable the achievement of the following \emph{desiderata}: extracting the minimum number of different predictive rules (one per cluster) having the lowest possible predictive error.
A number of challenges arise from the aforementioned workflow.
\begin{inlinelist}
	\item How to fix the correct number of clusters to identify?
	\item How to handle outliers in the construction of hypercubes around the found regions?
	\item How to build hypercubes around separable clusters associated with overlapping hypercubic regions?
\end{inlinelist}

We believe that powerful strategies to describe non-trivial clusters may exploit \emph{difference cubes} -- e.g., regions of the input feature space having a non-cubic shape and described by the subtraction of cubic areas -- and hierarchical clusters.
We stress that the importance of adopting hypercubes to describe input regions depends on the possibility to define a hypercube in terms of single variables belonging to specific intervals, in a highly human-comprehensible form.
This is not true when dealing with other representations---e.g., oblique rules, m-of-n rules.

\section{Conclusions}\label{sec:conclusions}

In this paper we propose a (deep) clustering-based workflow to enhance the symbolic knowledge extraction procedures from BB regressors in terms of computational complexity, fidelity and predictive performance.
Our method can be exploited to build hypercubic regions associated with human-readable logic rules in presence of linearly separable clusters of input instances.
In our future works we plan to implement and include in the \psyke{} framework~\cite{psyke-woa2021,psyke-ia2022,psyke-extraamas2022} different knowledge extractors adhering to the presented concepts and capable of handling more complex situations---e.g., outliers, clusters with more challenging shapes, non-linearly separable clusters.

\bibliographystyle{splncs04}
\bibliography{XLoKR-2022-clustering}

\end{document}